\def\year{2020}\relax
\documentclass[letterpaper]{article} 
\usepackage{aaai20}  
\usepackage{times}  
\usepackage{helvet} 
\usepackage{courier}  
\usepackage[hyphens]{url}  
\usepackage{graphicx} 
\usepackage{graphicx}  
\usepackage{ragged2e}
\usepackage{enumitem}
\usepackage{color}
\definecolor{guzcolor}{rgb}{0.0,0.2,0.8}
\definecolor{fcolor}{rgb}{0.3,0.6,0.1}

\title{Explainability via Responsibility}
\author {
Faraz Khadivpour and Matthew Guzdial\\
Department of Computing Science, Alberta Machine Intelligence Institute (Amii) \\
University of Alberta, Canada\\ \{khadivpour, guzdial\}@ualberta.ca \\
}
 
\begin{document}

\maketitle
\begin{abstract}

Procedural Content Generation via Machine Learning (PCGML) refers to a group of methods for creating game content (e.g. platformer levels, game maps, etc.) using machine learning models. 
PCGML approaches rely on black box models, which can be difficult to understand and debug by human designers who do not have expert knowledge about machine learning. 
This can be even more tricky in co-creative systems where human designers must interact with AI agents to generate game content. 
In this paper we present an approach to explainable artificial intelligence in which certain training instances are offered to human users as an explanation for the AI agent's actions during a co-creation process. 
We evaluate this approach by approximating its ability to provide human users with the explanations of AI agent's actions and helping them to more efficiently cooperate with the AI agent.

\end{abstract}

\section{Introduction}

In science and engineering, a black box is a component that cannot have its internal logic or design directly examined. 
In artificial intelligence (AI), ``The black box problem" refers to certain kinds of AI agents for which it is difficult or impossible to naively determine how they came to a particular decision \cite{zednik2019solving}.
Explainable artificial intelligence (XAI) is an assembly of methods and techniques to deal with the black box problem \cite{biran2017explanation}. Machine Learning (ML) is a subset of artificial intelligence that focuses on computer algorithms that automatically learn and improve through experience. \cite{goodfellow2016deep}. 
The current state-of-the-art models in ML, deep neural networks, are black box models.
Intuitively, it is difficult to cooperate with an individual when you cannot understand them. 
This is critical in co-creative systems (also called mixed-initiative systems), in which a human and an AI agent work together to produce the final output. \cite{yannakakis2014mixed}. 

There is a wealth of existing methods in the field of XAI \cite{adadi2018peeking}. 
For example, those that draw comparisons between the input and the output of a model \cite{cortez2011opening,cortez2013using,simonyan2013deep,bach2016controlling,dabkowski2017real,selvaraju2017grad}, or analyze the output in terms of the model's parameters \cite{boz2000converting,garcia2009enhancing,letham2015interpretable,hara2018making}. 
Alternatively, there is the strategy to attempt to simplify the model \cite{che2015distilling,tan2017detecting,xu2018interpreting}.
The major difference between our approach and these previous ones is that we present a method which makes it possible to explain an AI agent's action through a detailed inspection of what it has learned during the training phase.

Questions we might want to ask an AI agent include ``How did you learn to do that action?" or ``What did you learn that led you to make that decision?" \cite{cook2019framing}.
We sought to develop an approach that could answer these questions.
Thus, our approach needed to find explanations for the AI agent's decisions based on its training data. 

In this paper, we make use of the problem domain of a co-creative Super Mario Bros. level design agent. 
We use this domain since XAI is critical in co-creative systems. 
We introduce an approach to detect the training instance that is \emph{most responsible} for an AI agent's action. 
We can then present the most responsible training instance to the human user as an answer to how the AI agent learned to make a particular decision.  
To evaluate this approach we compare the quality of these responsible training instances to random instances as explanations in two experiments on existing data.


\section{Related Work}

Our problem domain is generating explanations for a PCGML co-creative agent. 
Therefore we separate the prior related work into three main areas: Procedural Content Generation via Machine Learning (PCGML), co-creative systems, and Explainable Artificial Intelligence (XAI).

\subsection{Procedural Content Generation via Machine Learning (PCGML)} 
Procedural Content Generation via Machine Learning (PCGML) is a field of research focused on the creation of game content by machine learning models that have been trained on existing game content \cite{summerville2018procedural}. 
Super Mario Bros. level design represents the most consistent area of research into PCGML. 
Researchers have applied many machine learning methods such as Markov chains \cite{snodgrass2016learning}, Monte-Carlo Tree Search (MCTS) \cite{summerville2015mcmcts}, Long Short-Term Recurrent Neural Networks (LSTMs) \cite{summerville2016super}, Autoencoders \cite{jain2016autoencoders}, Generative Adversarial Neural Networks (GANs) \cite{volz2018evolving}, and  genetic algorithms through learned evaluation functions \cite{dahlskog2014multi} to generate these levels. 
In a recent work, Khalifa et al proposed a framework to generate game levels using Reinforcement Learning (RL), though they did not evaluate it in Super Mario Bros. \cite{khalifa2020pcgrl}.
We also draw on reinforcement learning for our agent, however our approach differs from this prior work in terms of focusing on explainability.

\subsection{Co-creative systems}

There are numerous prior co-creative systems for game design. 
These approaches traditionally have not made use of ML, instead they rely on approaches like heuristics search, evolutionary algorithms, and grammars \cite{smith2010tanagra,liapis2013sentient,yannakakis2014mixed,deterding2017mixed,baldwin2017mixed,charity2020baba}. 
ML methods have only recently  been incorporated into co-creative game content generation.
Guzdial et al. proposed a Deep RL agent for co-creative Procedural Level Generation via Machine Learning (PLGML) \cite{guzdial2018co}. 
In another recent work, Schrum et al. presented a tool for applying interactive latent variable evolution to generative adversarial network models that produce video game levels \cite{schrum2020interactive}.
The major difference between our approach and previous ones is that it explains an AI partner's actions based on what it learned during training. 

It is important to note that we are not actually evaluating our approach in the context of co-creative interaction with a human subject study. We are only making use of data from prior studies in which humans interacted with ML and RL agents in co-creative systems.  

\subsection{Explainable Artificial Intelligence (XAI)}

The majority of existing XAI approaches can be separated according to which of two general methods they rely on: (\textbf{A}) visualizing the learned features of a model \cite{erhan2009visualizing,simonyan2013deep,nguyen2015deep,nguyen2016multifaceted,nguyen2017plug,olah2017feature,weidele2019deepling} and (\textbf{B}) demonstrating the relationship between neurons \cite{zeiler2014visualizing,fong2017interpretable,selvaraju2017grad}. 
Olah et al. developed a unified framework that included both (A) and (B) methods. \cite{olah2018building}. 

There are a few prior works focused on XAI applied to game design and game playing. 
Guzdial et al. presented an approach to Explainable PCGML via Design Patterns in which the design patterns act as a vocabulary and mode of interaction between user and model \cite{guzdial2018explainable}. 
Ehsan et al. introduced AI rationalization, an approach for explaining agent behavior for automated game playing based on how a human would explain a similar behavior \cite{ehsan2018rationalization}. 
Zhu et al. proposed a new research area of eXplainable AI for Designers (XAID) to help game designers better utilize AI and ML in their design tasks through co-creation \cite{zhu2018explainable}.

There exist a few approaches to explain RL agent's actions \cite{puiutta2020explainable}. 
Madmul et al. presented an approach that learns structural causal models to derive causal explanations of the behavior of model-free RL agents \cite{madumal2019explainable}. 
Kumar et al. presented a deep reinforcement learning approach to control an energy storage system. 
They visualized the learned policies of the RL agent through the course of training and visualized the strategies followed by the agent to users \cite{kumar2019explainable}.
Cruz et al. proposed a memory-based explainable reinforcement learning (MXRL) where an agent explained the reasons why some decisions were taken in certain situations using an episodic memory \cite{cruz2019memory}. 
In another recent paper, an approach was presented that employs explanations as feedback from humans in a human-in-the-loop reinforcement learning system~\cite{guan2020explanation}. 

To the best of our knowledge, this is the first XAI work focused on the training data of a target ML model.
Our approach differs from existing XAI work in detailed inspection and alteration of the training phase.

\section{System Overview}

\begin{figure}[tb]
  \includegraphics[width=8.3cm,height=3.5cm]{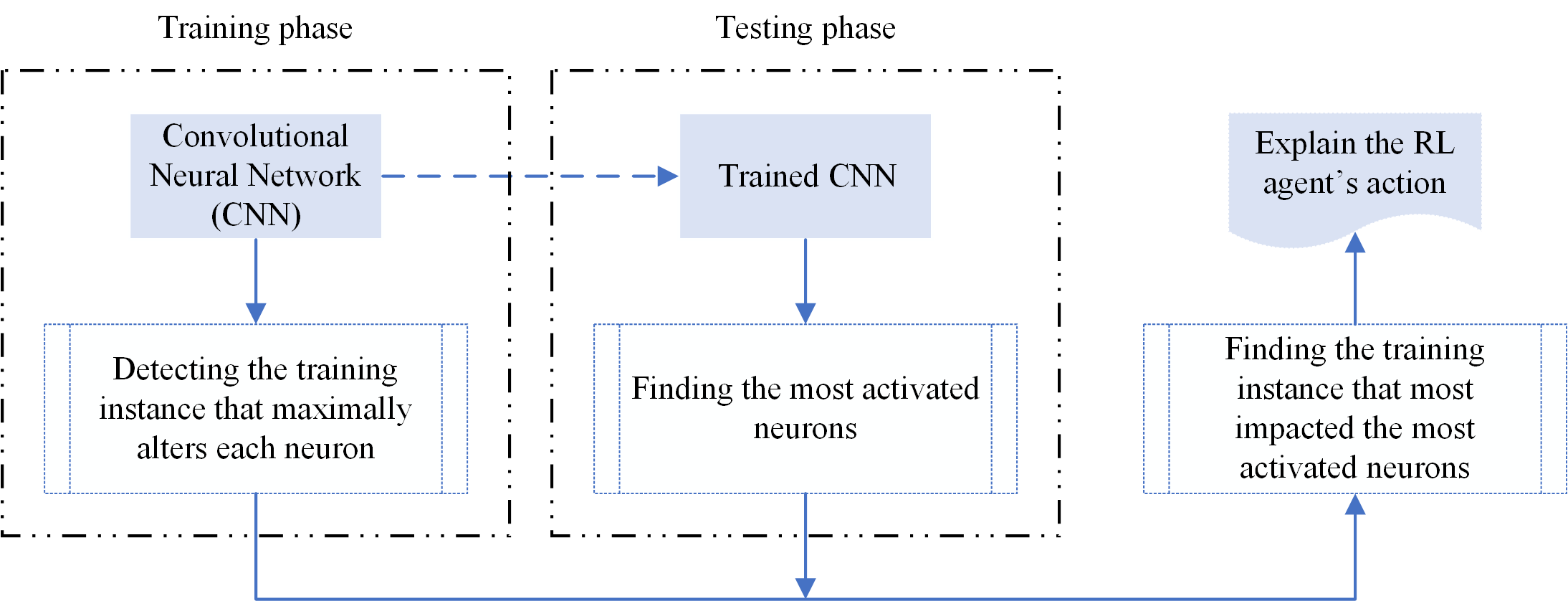}
  \caption{General steps of our approach}
  \label{fig:1}
\end{figure}

\begin{figure}[tb]
  \includegraphics[width=8.5cm,height=5cm]{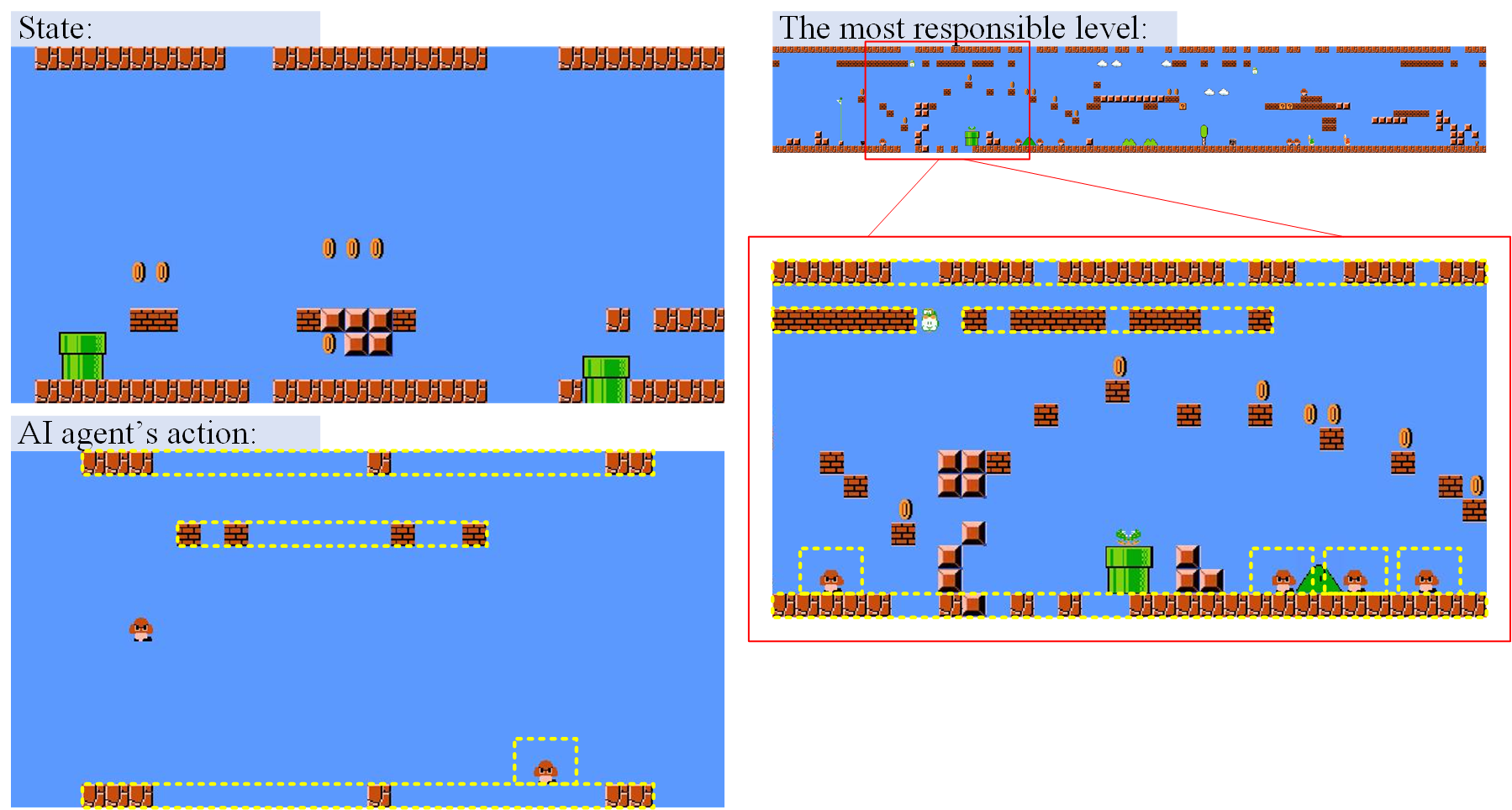}
  \caption{An example of explaining an AI agent's action by representing the most responsible level.}
  \label{fig:3}
\end{figure}

In this paper, we present an approach for Explainable AI (XAI) that aims to answer the question ``What did the AI agent learn during training that led it to make that specific action?".
As is shown in Figure \ref{fig:1}, the general steps of the approach are as follows: First, during training a DNN, we detect the training instance (or instances) that maximally alters each neuron inside the network. 
Secondly, during testing, we pass each instance through the network and find the neuron that is most activated \cite{erhan2010understanding}. 
Then given the information from the first step, we can easily identify an instance (or instances) from the training data that maximally impacted the most activated neuron. 
We refer to this as ``\emph{the most responsible training instance}" for the AI agent's action.
The intuition is that the user can take this explanation as something akin to the end goal of the agent taking that action.
Our hope is that it will be helpful in the user deciding whether to keep or remove some addition by the AI.
For example in Figure \ref{fig:3}, given the most responsible level as the explanation, the user might keep the lower of the two Goombas, despite the fact that it seems to be floating, if they can match it to the Goombas from the most responsible level.

For this purpose, we pre-trained a Deep RL agent using data from interactions of human users with three different ML level design partners (LSTM, Markov Chain, and Bayes Net) to generate the Super Mario Bros level.
This is the same Deep RL architecture and data from prior work by Guzdial et al. \cite{guzdial2018co} for co-creative Procedural Level Generation via Machine Learning (PLGML), in which they made use of the level design editor from \cite{guzdial2017general} which is publicly online.\footnote{1https://github.com/mguzdial3/Morai-Maker-Engine}
The agent is designed to take in a current level design state and to output additions to that level design, in order to iteratively complete a level with a human partner.

\begin{figure}[tb]
  \includegraphics[width=8.5cm,height=4cm]{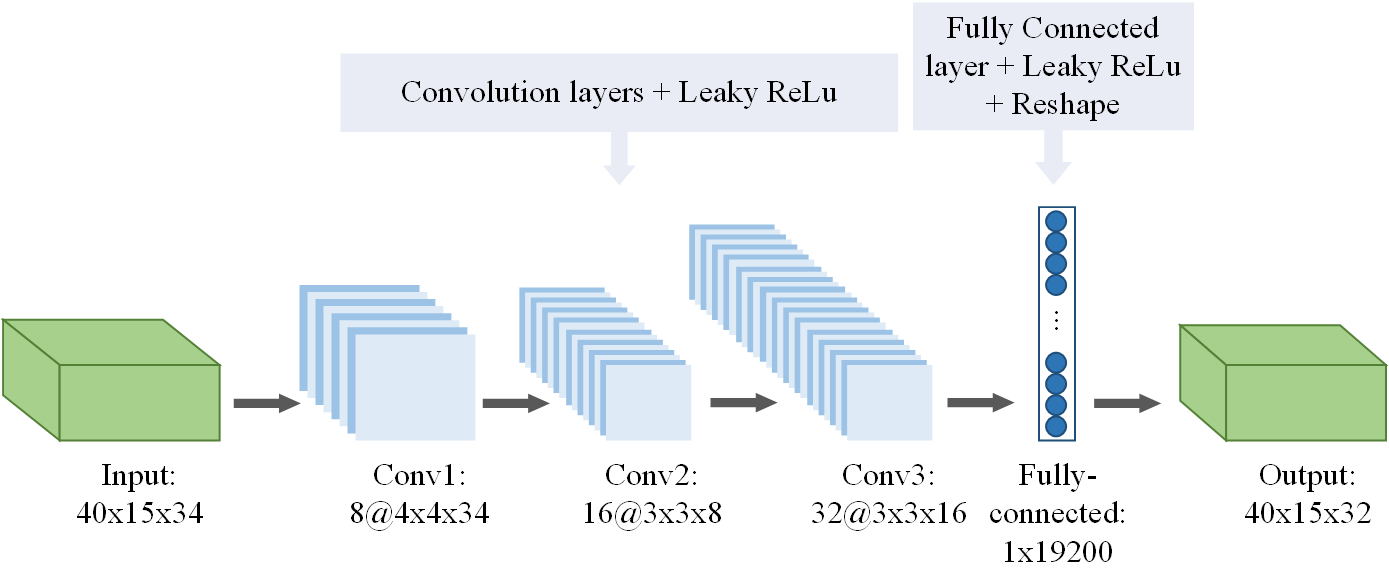}
  \caption{Architecture of our Convolutional Neural Network (CNN).}
  \label{fig:2}
\end{figure}

Our training inputs are states and the outputs are the Q table values for taking a particular action for the particular state. 
The input comes into the network as a state of shape (40x15x34). 
The 40 is the width and 15 is the height of a level chunk. 
At each x,y location there are 34 possible level components (e.g. ground, goomba, pipe, mushroom, tree, Mario, flag, ...) that could be placed there. 
As is shown in the visualized architecture of the Convolutional Neural Network (CNN) in Figure \ref{fig:2}, it has three convolutional layers and a fully connected layer followed by a reshaping function to make the output in the form of the action matrix which is (40x15x32). 
The player (Mario) and flag are the level entities that cannot be counted as an action, so there are 32 possible action components instead of the 34 state entities.
Our activation function is ``Leaky ReLu" for every layer and the loss function is ``Mean Squared Error" and the optimizer is ``Adam", with the network built in Tensorflow \cite{abadi2016tensorflow}. 
We make use of this existing agent and data since it is the only example of a co-creative PCGML agent where the data from a human subject study is publicly available.

During each training epoch we employ a batch size of one to track when each training instance passes through the network. 
We calculate and store the change of neuron weights between batches. 
After training, by summing over the changes of each neuron weight with respect to training data, we are able to identify which training instance maximally results in alteration of a neuron.
Since positive and negative values can counteract each other's effects, it is important to not look at the absolute values until the end of the training. 
We can then sum and store this information inside eight arrays of shape (4x4x34) for the first convolutional layer, 16 arrays of shape (3x3x8) for the second convolutional layer, and 32 arrays of shape (3x3x16) for the third convolutional layer.
These are the shapes of the filters in each layer. 
We name these arrays Most Responsible Instance for each Neuron in each Convolutional layer (MRIN-Conv1, MRIN-Conv2, and MRIN-Conv3). 
These data representations link neurons to IDs representing a particular instance of a human user working with the AI in the co-creative tool. 
We can then search these arrays and find the ID of a training instance that is the most responsible for changes to a particular weight.

Our end goal is to determine the most responsible training instance for a particular prediction made by our trained CNN. 
To do that, we need to find out what part of the network was most important in making that prediction. 
We can then determine the most responsible instance for the final weights of this most important part of the network. 
The most activated filter of each convolutional layer is a filter that contributes to the slice with the largest magnitude in the output of that layer. 
Hence the most activated filter can be considered the most important part of the convolutional layer for that specific test instance \cite{erhan2010understanding}. 
For example, we pass a test instance into the network. 
A test instance is a (40x15x34) state that is a chunk of a partially designed level. Since the first convolutional layer has 8 4x4x34 filters with the same padding, the output would be in the shape of (40x15x8).
Then we find the (40x15) slice with the largest values. 
The most activated filter is a (4x4x34) array in our convolutional layer which led to the slice with the greatest magnitude. 

Finally, once we have the maximally activated filter we can identify \emph{the most responsible} training instance (or instances) by querying the MRIN-Conv arrays we built during training. 
The most responsible training instance is the ID that most repeated in the MRIN-Conv array associated with the maximally activated filter. 
We chose the most repeated ID since it is the one that most frequently impacted the majority of the neurons in the filter during training.

\section{Evaluation}

In this section, we present two evaluations of our system. 
We call the first evaluation our ``Explainability Evaluation'' as it addresses the ability of our system to provide explanations that help a user  predict an AI agent's actions. 
We call the second evaluation our ``User Labeling Error Evaluation'' as it addresses the ability of our system to help human users identify positive and negative AI additions during the co-creative process. 
Both evaluations approximate the impact of our approach on human partners by using existing data of AI-human interactions.
Essentially, we act as though the pre-recorded actions of the AI agent were outputs from our Deep RL agent and identify the responsible training instances as if this were the case.
Due to the fact that our system derives examples as explanations for the behavior of a co-creative Deep RL agent, a human subject study would be the natural way to evaluate our system. 
However, prior to a human subject study, we first wanted to gather some evidence of the value of this approach.

\subsection{Explainability Evaluation}

The first claim we made was that this approach can help human users better understand and predict the actions of an AI agent. 
In this experiment we use the most responsible level as an approximation of the AI agent's goal, in other words what final level the AI agent is working towards. 
The most responsible level refers to a level at the end of a human user's interactions with an AI agent.
We identify this level by finding the most responsible training instance as above and identifying the level at the end of that training sequence.
This experiment is meant to determine if this can help a user to predict the AI agent's actions.
To do this, we passed test instances into our network and found the most responsible training instances. 
We then compared the most responsible level for some current test instance to the AI agent's action in the next test instance. 
If the most responsible level is similar to the action it would indicate that the most responsible level can be a potential explanation for the AI agent's action by priming the user to better predict future actions by the AI agent.
In comparison, we randomly selected 20 levels from the training data and found their similarities to the AI agent's action in the next test instance. 
If our approach outperforms the random levels, it will support the claim that the responsible level is better suited to helping predict future AI agent actions compared to random levels.

We used two different sets of test data: 
\begin{enumerate}[label=(\Alph*)]
\item Our first testset is derived from a study in which users interacted with pairs of three different ML agents as mentioned in our System Overview section \cite{guzdial2018co}. 
We used the same testset identified in that paper.
\item Our second testset is obtained from a study in which expert level designer users interacted with the trained Deep RL agent \cite{guzdial2019friend}.
\end{enumerate}

If we find success with the first testset then that would indicate that our trained Deep RL agent is a good surrogate for the original three ML agents, since we would be in effect predicting the next action of one of these agents.
Good results for the second testset would demonstrate the capability for prediction of the Deep RL agent's actions itself. 
Since the first convolutional layer is the layer that most directly reasons over the level structure, we decided to find the most responsible training instance of just the first convolutional layer. 
However, this setup puts our approach at a disadvantage, since we are going to compare only one most responsible level to 20 random ones.

For comparing the most responsible level and the random levels to the actions, we needed to define a suitable metric. 
We desired a metric that detects local overlaps and represents the similarity between a level and action.
We wanted to pick square windows which are not the same size as the first convolutional layer, to capture some local structures without biasing the metric too far towards our first convolutional layer. 
As a result, we found all three-by-three non-empty patches for both a given level and an action.
Then we counted the number of exact matches of these patches on both sides, removing the matched ones from the dataset since we wanted to count the same patches only once.
Finally, we divided the total number of the matched patches by the total number of patches in the action, since this was always smaller than the number from the level. 
We refer to this metric as the \emph{local overlap ratio}.

\subsection{Explainability Evaluation Results}

We had 242 samples in the first testset and 69 samples in the second one. 
Since we wanted to compare instances in which the AI agent actually made some serious changes, we chose instances where the AI agent added more than 10 components in its next action. 
Thus we came to 38 and 46 instances from the first and second testsets, respectively.

Our approach outperforms the random baseline in 78.94 percent of 38 instances for the ML agents data and 67.29 percent of 46 instances for the Deep RL agent data. 
The average of the local overlap ratios is shown in Table \ref{table:1} (higher is better). 
The minimum value here would be 0 for zero overlap and the maximum value would be 1 for complete overlap between the action and the most responsible level or the random level. 
This normalization means that even small differences in this metric represent large perceptual differences.
For example, a 0.04 difference in the local overlap ratio between the most responsible level and the random levels in Table \ref{table:1} indicates the most responsible level has 20 more three-by-three non-empty overlaps.
We expect that the reason that the Deep RL agent values are generally lower is that the second study made use of published level designers rather than novices and an adaptive Deep RL Agent, meaning that there was more varied behavior compared with the three ML agents. 

\begin{table}[t]
\begin{tabular}{||c||c|c||} 
\hline
TestSet & Most Responsible Level & Random Levels \\  
\hline\hline
ML Agents & \textbf{0.4653} & 0.3841 \\
\hline
Deep RL & \textbf{0.2880} & 0.2472 \\
\hline
\end{tabular}
\caption{A table comparing the average of the most responsible levels to the average of the random levels for both testsets.}
\label{table:1}
\end{table}

An example of explainability is demonstrated in Figure \ref{fig:3}. As is shown in the figure, the AI agent made an action and added some components (e.g. goomba and ground) to the existing state. By looking at the chunk of the most responsible level, the user might realize that the AI agent wants to generate a level including some goombas as enemies and some blocks in the middle of the screen. 
The AI agent also added ground at the bottom and top of the screen, which the user could identify as being consistent with both their input to the agent and the most responsible level.

\subsection{User Labeling Error Evaluation}

\begin{figure*}[t]
 \begin{center}
    \includegraphics[width=16cm,height=8cm]{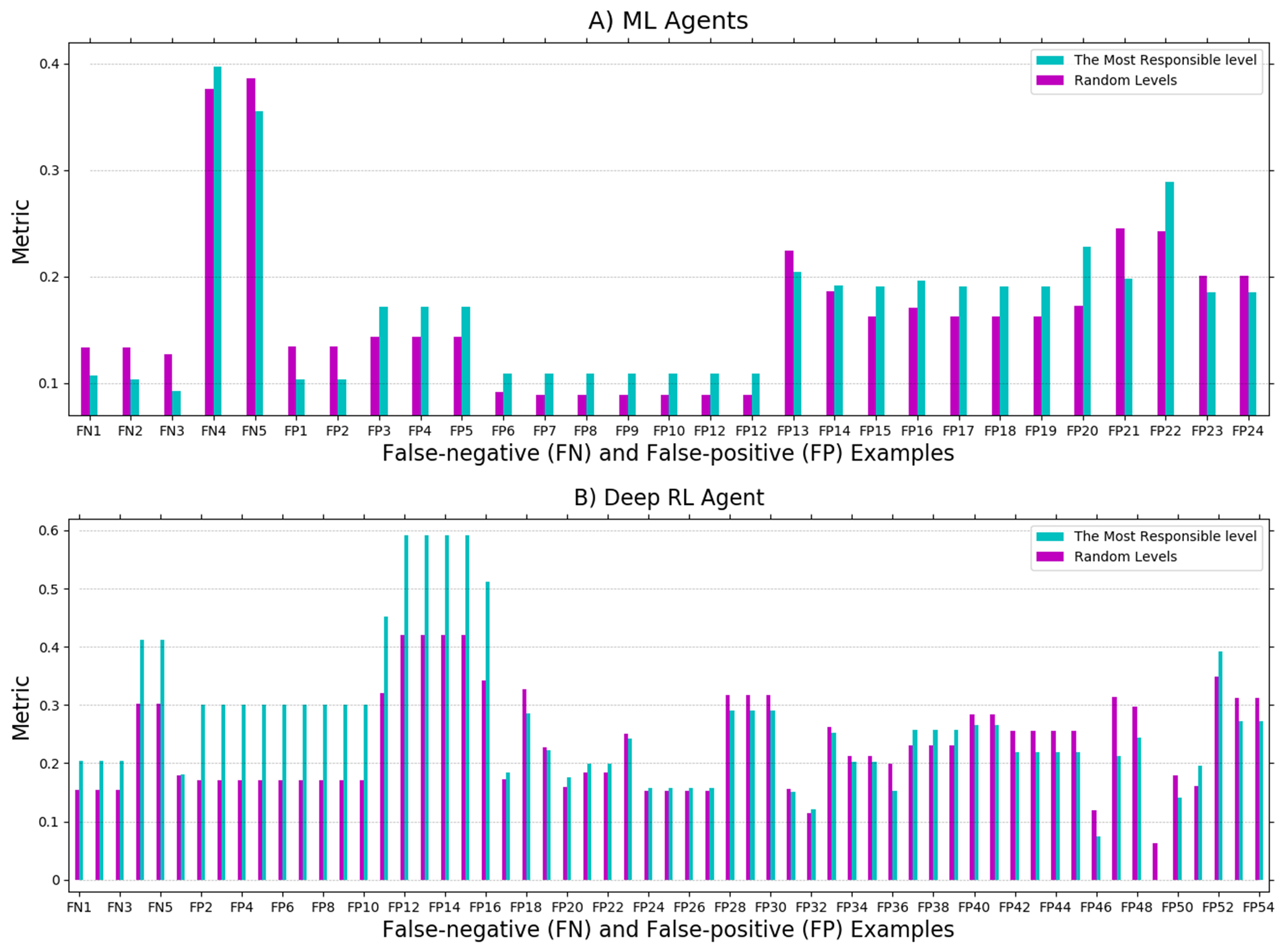}
 \caption{Results of the User Labeling Error Evaluation.} 
 \label{fig:4} 
 \end{center}
\end{figure*}

For the second evaluation, we wanted to get some sense of whether this approach could be successful in terms of assisting a human user in better understanding good and bad agent actions during the co-creation process. 
To do this, we needed to identify specific instances where our tool could be helpful in the data we have available.
We defined two such concepts: (A) false-positive decisions and (B) false-negative decisions, based on the interactions between users and AI partner during level generation:
\begin{enumerate}[label=(\Alph*)]
\item \textbf{False-positive decisions} are additions by the AI partner that the user kept at first but then deleted later. 
\item \textbf{False-negative decisions} are additions by the AI partner that the user deleted at first but then added later.
\end{enumerate}
Given these concepts, if we could help the user avoid making these kinds of decisions, our approach could help a human user during level generation.
We anticipated that one reason that users made these kinds of decisions was from a lack of context of the AI agent's action. 
Thus, if the user had context they may not delete or keep what they would otherwise keep or delete, respectively.

To accomplish this, we implemented an algorithmic way to determine false-positives and false-negatives among the two testsets described in the previous evaluation. 
In this algorithm, we first find all user decisions in terms of deleting or keeping an addition by the AI agent. 
Then we look at the level at the end of the user and the AI agent's interaction. 
If a deleted AI addition exists in the final level, it is counted as a false-negative example, and if a kept addition does not exist in the final level it is counted as a false-positive example. 

Once we discovered all false-negative and false-positive examples, we found the state before the example was added by the AI agent and named it the Introduction-state (I-state). 
We found the state in which false-positivity or false-negativity occurred (i.e. when a user re-added a false-negative or deleted a false-positive) and named it the Contradiction-state (C-state). 
Since some change between the I-state and the C-state led to the user altering their decision, we wanted to see some sign that presenting the most responsible level to the user could change their mind before they reached this point. 
Thus we compared these two states to find all the changes that the AI agent or the user made and named this the Difference-state (D-state).

We compared each D-state with the final generated level derived from the most responsible training instance. 
We also compared each D-state with 20 other randomly selected levels from the existing data.
For the comparison, we used the local overlap ratio defined in the previous evaluation. If our approach outperforms the random baseline, we will be able to say that there is some support for the responsible level helping the user avoid false-positives and false-negatives in comparison to random levels.

\subsection{User Labeling Error Evaluation Results}

We found five false-negative and 24 false-positive examples in the first testset and five false-negative and 54 false-positive examples in the second one. 
The results of the evaluation are demonstrated in Figures \ref{fig:4}.

For the first dataset which included the actions of the three ML agents, our approach outperformed the random baseline in 65.51 percent of the examples. 
The average of the local overlap ratio values for our approach was 0.1717 which is more than the 0.1647 for the random levels. 
For the second dataset obtained from the Deep RL agent, our approach outperformed the baseline in 59.32 percent of the examples. 
The average of the local overlap ratio values were 0.2665 and 0.2328 for the most responsible level and random levels, respectively. 
Again this represents a large perceptual difference of roughly 15 more non-empty 3x3 overlaps.  

Interestingly, our approach outperforms the random levels in all of the false-negative examples in the second dataset, compared with just 20 percent of false-negatives in the first dataset. 
Further, our approach performs around 1.5 times better than the random levels in 15 false-positive examples in the second dataset.
These instances come from the study that used the same RL agent as we used to derive our explanations, which could account for this performance.

\section{Discussion}

In this paper, we present an XAI approach for a pre-trained Deep RL agent.
Our hypothesis was that our method could be helpful to human users. 
We evaluated it by approximating this process for two tasks using two existing datasets. 
These datasets are obtained from studies using three ML partners and an RL agent. 
Essentially, we used the XAI-enabled agent in this paper as if it were the agents used in these datasets.
The results of our first evaluation demonstrates that our method is able to represent examples as explanations to help users predict an agent's next action.
The results of our second evaluation support our hypothesis and give us an initial signal that this approach could be successful in order to help human users more efficiently cooperate with a Deep RL agent. 
This indicates the ability of our approach to help human designers by presenting an explanation for an AI agent's actions during a co-creation process. 

A human subject study would be a more reasonable way to evaluate this system since human users might be able to derive meaning from the responsible level that our similarity metric could not capture. 
Our approach performs better than our baseline of random levels in both evaluation methods and this presents evidence towards its value at this task. 
However, we look forward to investigating a human subject study in order to fully validate these results.

There could be other alternatives to a human subject study. For example, a secondary AI agent that predicts our primary AI agent's actions can play a human partner's role in the co-creative system. 
Thus making use of a secondary AI agent to evaluate our system before running a human subject study might be a simple next step. 

It is important to mention that we only offer one most responsible level from only the first convolutional layer as an explanation. 
Looking into providing a user with multiple responsible levels or looking into the most responsible levels of the other layers could be a potential way to further improve our approach. Our metric for determining the most responsible training instance is based on finding the most repeated instance inside the MRIN-Conv arrays associated with the most activated filter. 
We identified the most activated filter by looking at the absolute values.
We plan to investigate other metrics such as looking for the most activated neurons outside of the filters.
In addition, considering negative and positive values separately in the maximal activation process could also lead to improved behavior. Negative values might indicate that an instance negatively impacted a neuron. It could be the case then that the filter might be maximally activated because it was giving a very strong signal against some action. 

One quirk of our current approach is that the most responsible training instance depends on the order in which it was presented to the model during the training.
Thus, this measure does not tell us about any inherent quality of a particular training data instance, only it's relevance to a particular model that has undergone a particular training regimen. 
In the future, we intend to explore how more general representations of responsibility such as Shapely values might intersect with this approach \cite{ghorbani2019data}.

Only the domain of a co-creative system for designing Super Mario Bros. levels is explored in this paper.
Thus making use of other games will be required to ensure this is a general method for level design co-creativity. 
Beyond that, we anticipate a need to demonstrate our approach on different domains outside of games.
We look forward to running another study to apply our approach to human-in-the-loop reinforcement learning or other co-creative domains.

\section{Conclusions}

In this paper we present an approach to XAI that provides human users with the most responsible training instance as an explanation for an AI agent's action. 
In support of this approach, we present results from two evaluations. 
The first evaluation demonstrates the ability of our approach to offer explanations and to help a human partner predict an AI agent's actions.
The second evaluation demonstrates the ability of our approach to help human users better identify good and bad instances of an AI agent's behavior.
To the best of our knowledge this represents the first XAI approach focused on training instances.

\section*{Acknowledgements}

We acknowledge the support of the Natural Sciences and Engineering Research Council of Canada (NSERC) and the Alberta Machine Intelligence Institute (Amii). 

\bibliographystyle{aaai.bst}
\bibliography{main.bib}

\end{document}